\documentclass[acmtog]{acmart}
\begin{document}
\newcommand{\bm}[1]{\mbox{\boldmath{$#1$}}}
\newcommand{\figlabel}[1]{\label{fig:#1}}
\newcommand{\tablabel}[1]{\label{tab:#1}}
\newcommand{\equlabel}[1]{\label{equ:#1}}
\newcommand{\figref}[1]{Fig.~\ref{fig:#1}}
\newcommand{\tabref}[1]{Table~\ref{tab:#1}}
\newcommand{\equref}[1]{(\ref{equ:#1})}
\newcommand{\alglabel}[1]{\label{alg:#1}}
\newcommand{\algref}[1]{{\bf Algorithm}~\ref{alg:#1}}
\newcommand{\seclabel}[1]{\label{sec:#1}}
\newcommand{\secref}[1]{Section~\ref{sec:#1}}

\title{A General Scattering Phase Function for Inverse Rendering}

\author{Thanh-Trung Ngo}
\authornote{}
\email{trung@gmail.com}
\orcid{}
\author{Hajime Nagahara}
\authornotemark[1]
\email{nagahara@ids.osaka-u.ac.jp}
\affiliation{%
  \institution{Institute for Datability Science, Osaka University}
  \streetaddress{}
  \city{}
  \state{}
  \country{}
  \postcode{}
}

\begin{abstract}
  We tackle the problem of modeling the light scattering in homogeneous translucent material and estimating its scattering parameters. A scattering phase function is one of such parameters which affects the distribution of scattered radiation. It is the most complex and challenging parameter to be modeled in practice, and empirical phase functions are usually used. Empirical phase functions (such as Henyey-Greenstein (HG) phase function or its modified ones) are usually presented and limited to a specific range of scattering materials. This limitation raises concern for an inverse rendering problem where the target material is generally unknown. In such a situation, a more general phase function is preferred. Although there exists such a general phase function in the polynomial form using a basis such as Legendre polynomials~\cite{Fowler1983}, inverse rendering with this phase function is not straightforward. This is because the base polynomials may be negative somewhere, while a phase function cannot. This research presents a novel general phase function that can avoid this issue and an inverse rendering application using this phase function. The proposed phase function was positively evaluated with a wide range of materials modeled with Mie scattering theory. The scattering parameters estimation with the proposed phase function was evaluated with simulation and real-world experiments.
\end{abstract}


\begin{teaserfigure}
  \includegraphics[width=\textwidth]{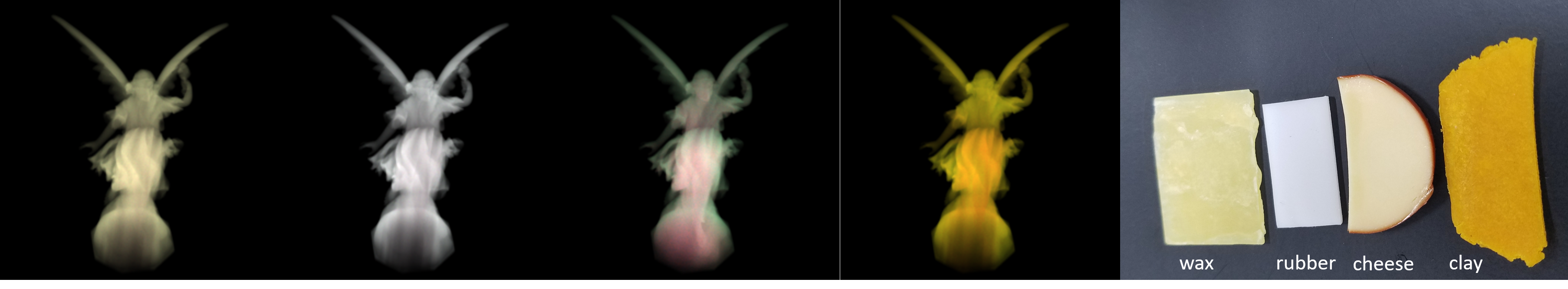}
  \caption{The renders of lucy with estimated scattering parameters for the actual material slabs on the right. The scattering parameters are estimated and rendered independently for all the color channels before combining them to make the color images. }
  \figlabel{teaser}
\end{teaserfigure}

\maketitle

\section{Introduction}
The scattering of light is a physical phenomenon seen in participating media and scattering material. It has been using to study the galaxy~\cite{Henyey1941,Witt_Nature_1968}, the earth's atmosphere~\cite{Rayleigh1899,Hansen_1974}, biological tissues~\cite{Saidi_95,Cheong_biologicaltissues_1990,Bhandari_11,Yang_14}, seawater~\cite{Zhang_19}, optical properties of translucent material~\cite{Frisvad2020} and material fabrication~\cite{Papas2013}. Scattering of light in the participating media is usually modeled by the radiative transfer equation (RTE)~\cite{chandrasekhar1950radiative} with three important parameters (scattering coefficient, absorption coefficient, and a phase function), where the phase function is the most complex and challenging to be formulated mathematically. Although the phase function can be theoretically formulated by Mie scattering theory~\cite{Pierrehumbert2011}, it is hard to use this theory in practice because it requires the size of the scatterers and the shape must be approximated as a sphere or spheroid. In practice, the shape of scatterers is irregular and variable in size, such as dust, marine scatters, and blood cells. Therefore an empirical phase function is usually used in such a case.

Plenty of empirical phase functions have been presented for a specific range of scattering media. For examples, phase function was proposed to model galaxy dust~\cite{Henyey1941}, biomedical media~\cite{Liu_1994}, seawater~\cite{Haltrin_88,Fournier_1999}, planetary regoliths~\cite{Hapke1981}. However, if the media or material is unknown, we do not know which phase function is the best to model the light scattering. This is the case for inverse rendering. In such a case, a more general phase function is preferred.

There exist more general phase functions. The most widely used phase function can be HG phase function. Although it was originally proposed for studying galaxy dust, it is found to be useful in a wide range of applications mainly due to its simplicity. It is so popular that it has been extended to be more applicable and accurate in a wider range of applications. For examples, it is generalized using Gegenbauer polynomials~\cite{Raynold_Gen_HG_1980} or generalized along with other phase functions~\cite{Draine2003}. It is extended to have both forward and backward scattering lopes~\cite{KATTAWAR1975} or three lopes~\cite{Hong_hg3_1985} to better describe the scattering with both large and small scatterers. It is combined with the Rayleigh phase function~\cite{Rayleigh1899} to also better describe the scattering with small scatterers~\cite{Liu_06} with only a single parameter. It is also combined with additional different terms~\cite{Jacques_HGiso_2987,Fishburne1976Report} to increase its representation capability. There exist also extensions of HG phase function extension~\cite{Wang_pf2019,Baes2022}. However, these functions still inherit the limitations of the original HG phase function and are limited to some specific range of media and material. Fortunately, there exists a way that can theoretically model a more general phase function, such as using Legendre polynomials~\cite{Howell2020}, Gegenbauer polynomials~\cite{Raynold_Gen_HG_1980}, Taylor series~\cite{Sharma1998}. This approach can be used to represent any existing phase function. However, it is not straightforward to use a polynomial function using the above basis to represent a phase function even when its total probability density is normalized. The problem is that this type of function can exhibit a negative value while the phase function cannot. Of course, we can regularize the function to be positive in any observation angle, but it will reduce the representation capability and consume much computation power. This problem emerges when we solve the inverse rendering with unknown participating media. 

This research first presents a general scattering phase function that solves the aforementioned problem. The proposed phase function is theoretically guaranteed to fit any smooth positive phase function. Second, we present an application using the proposed phase function to estimate translucent material scattering parameters.

\section{Related Works}
\subsection{General Scattering Phase Function}
Besides the HG and its extended phase functions mentioned in the previous section, several general phase functions have been presented. Gkioulekas et al.\cite{Gkioulekas_SIGG_2013} presented a piece-wise linear phase function that linearly combines a dictionary or basis of 200 tent functions. This phase function can approximate any phase function. However, it requires too many parameters. Moreover, to estimate a phase function using this representation, a regularization to avoid high-frequency artifacts needs to be employed that hinders the estimation of highly forward/backward scattering phase functions. This phase function is not much different from a non-parametric phase function presented as a look-up-table phase function~\cite{Minetomo_2017}. An example of a polynomial phase function is presented by Sharma et al.~\cite{Sharma1998, Kokhanovsky_book2015}. These phase functions are used to represent the existing ones, such as the Mie phase functions.

\subsection{Scattering Parameter Estimation with Differential Rendering}
Inverse rendering with physics-based differentiable rendering~\cite{Zhao_SIGG2020} currently has become attractive in computer vision and computer graphics because it can be applied to a wide range of applications, including understanding a translucent material~\cite{Che_ICCP2020}. Employing a physics-based differentiable renderer, we can estimate the scattering parameters of the media easily without tricky settings of prior works~\cite{Narasimhan2006,Mukaigawa_CVPR2010,Minetomo_2017}.

\section{Exponential Phase Function}

\subsection{Forward Representation of Phase Function with Polynomials}
A scattering phase function is a parameter of the participating media that influences the scattering angular distribution in the 3-dimensional space of the light when interacting with the media particles. Because the distribution is symmetric about the incoming direction, a practical phase function is usually described as a 1-dimensional function $p(\theta)$ or $p(\cos\theta)$, where $\theta$ is the relative angle between an existing light direction and the incoming light direction.

Any smooth phase function can be approximated with a polynomial basis such as Legendre polynomials~\cite{Fowler1983,Howell2020}, Gegenbauer polynomials~\cite{Raynold_Gen_HG_1980}, or Taylor series\cite{Sharma1998}. In these approaches, a phase function can be represented by a linear combination of base polynomials:
    \begin{equation}
        p(\mu)=\sum_{i=0}^{N=\infty}a_i P_i(\mu),
        \equlabel{pf_poly}
    \end{equation}
    where $P_i(\mu)$ is a base polynomial, $\mu=\cos\theta$, and $a_i$ is a coefficient. $N$ defines the degree of the target polynomial. In practice we only need a small number of $N$ for an acceptable approximation accuracy. Further, the parameter of this polynomial phase function is normalized so that 
    \begin{equation}
        2\pi \int_{-1}^{1}p(\mu)=1.
        \equlabel{pf_norm}
    \end{equation}
    For instance, the HG phase function can be represented using Legendre polynomials~\cite{Kokhanovsky_book2015} as follows:
    \begin{equation}
        p_{HG}(\mu)=\sum_{i=0}^{\infty}g^i(2i+1)P_i(\mu),
    \end{equation}
    where $P_i(\mu)$ is a Legendre polynomial and $g$ is the asymmetry of the phase function that is defined by:
    \begin{equation}
        g=\int_{-1}^{1}\mu p(\mu) d\mu.
        \equlabel{g}
    \end{equation}
    
This presentation is guaranteed if we know the phase function beforehand. On the contrary, any function represented by~\equref{pf_poly} that is normalized by~\equref{pf_norm} is not guaranteed to be a phase function. This problem arises when we solve inverse rendering with a radiative transfer equation for participating media. We need an additional constraint on the parameters $\{a_i\}$ to constrain this polynomial as a phase function. Specifically, along with~\equref{pf_norm}, we need a constraint: 
    \begin{equation}
        p(\mu)>0~\forall \mu \in [-1,1]. 
        \equlabel{pf_positive_constraint}
    \end{equation}
However, this is a cumbersome constraint because we cannot afford to check for all possible $\mu$ exhaustively. Moreover, if we can do this, we have to pay a high computation cost. The constraint also limits the representation capability of the polynomial function and splits the coefficient searching space, thus introducing more local minima and hindering the inverse rendering optimization process. The following section presents a simple solution that relaxes this constraint.

\subsection{Exponential Phase Function}
A smooth and positive phase function $p(\mu)$ is in fact a probability density function with $0< p(\mu) <+\infty$. If we consider logarithm of this phase function we have $-\infty < \log(p(\mu))< +\infty$. Using function approximation methods that apply to the original phase function as described above, we can theoretically represent $\log(p(\mu))$ with a polynomial:
    \begin{equation}
        \log(p(\mu))=\sum_{i=0}^{M=\infty}b_i Q_i(\mu),
        \equlabel{epf_poly}
    \end{equation}
    where $Q_i$ is a base polynomial such as a Legendre polynomial. The advantage of this approach is that we can relax the constraint \equref{pf_positive_constraint} for $ \log(p(\mu))$. The only necessary constraint is the normalization \equref{pf_norm}.
    Further, the \equref{epf_poly} can be reformulated:
    \begin{equation}
        p(\mu)=\exp{\sum_{i=0}^{N=\infty}b_i Q_i(\mu)},
    \end{equation}
    and we call this an exponential phase function. This representation of a phase function clearly solves the aforementioned limitation of a general polynomial phase function. Although the $M$-degree exponential phase function has $M+1$ coefficients, $b_0$ is fixed by the normalization constraint~\equref{pf_norm}. Therefore, $M$-degree phase function has $M$ parameters.
    
    There are some special case of the proposed phase function. First, when we use zero-degree polynomial, $b_i=0$ for $i>0$, the phase function is constant. It is a isometric phase function. Second, when we use the one-degree polynomial, $b_1 \ne 0$ and $b_i=0$ for $i>1$, the phase function corresponds to the von Mises-Fisher distribution phase function (vMF) that is used by Gkioulekas et al.~\cite{Gkioulekas_pf_2013}:
    \begin{equation}
        p_{vMF}(\mu)=\frac{\kappa}{2\pi \sinh(\kappa)} \exp(\kappa\mu),
    \end{equation}
    where $\kappa$ is the only shape parameter. This phase function can be reformulated to have the same form of our exponential phase function:
    \begin{equation}
        p_{vMF} = \exp(\alpha+ \kappa\mu),
    \end{equation}
    where $\alpha$ is set so that $\exp(\alpha)=\frac{\kappa}{2\pi \sinh(\kappa})$. Third, when we use second-degree polynomial, $b_1 \ne 0$, $b_2 \ne 0$, and $b_i=0$ for $i>2$, the proposed phase function resembles the Fisher-Bingham distribution~\cite{Kent_dist_1982}. The examples of the proposed phase functions of zero-degree, one-degree, and two-degree polynomial representations are shown in~\figref{pf_examples}. For the two-degree phase function, the phase function is in fact the one that fits to Rayleigh phase function with $(b_1,b_2)=(0,0.68)$ and $b_0$ is fixed by the normalization constraint~\equref{pf_norm}. The examples show the usefulness of the proposed phase function even with a low degree.
    \begin{figure}
        \includegraphics[width=\columnwidth]{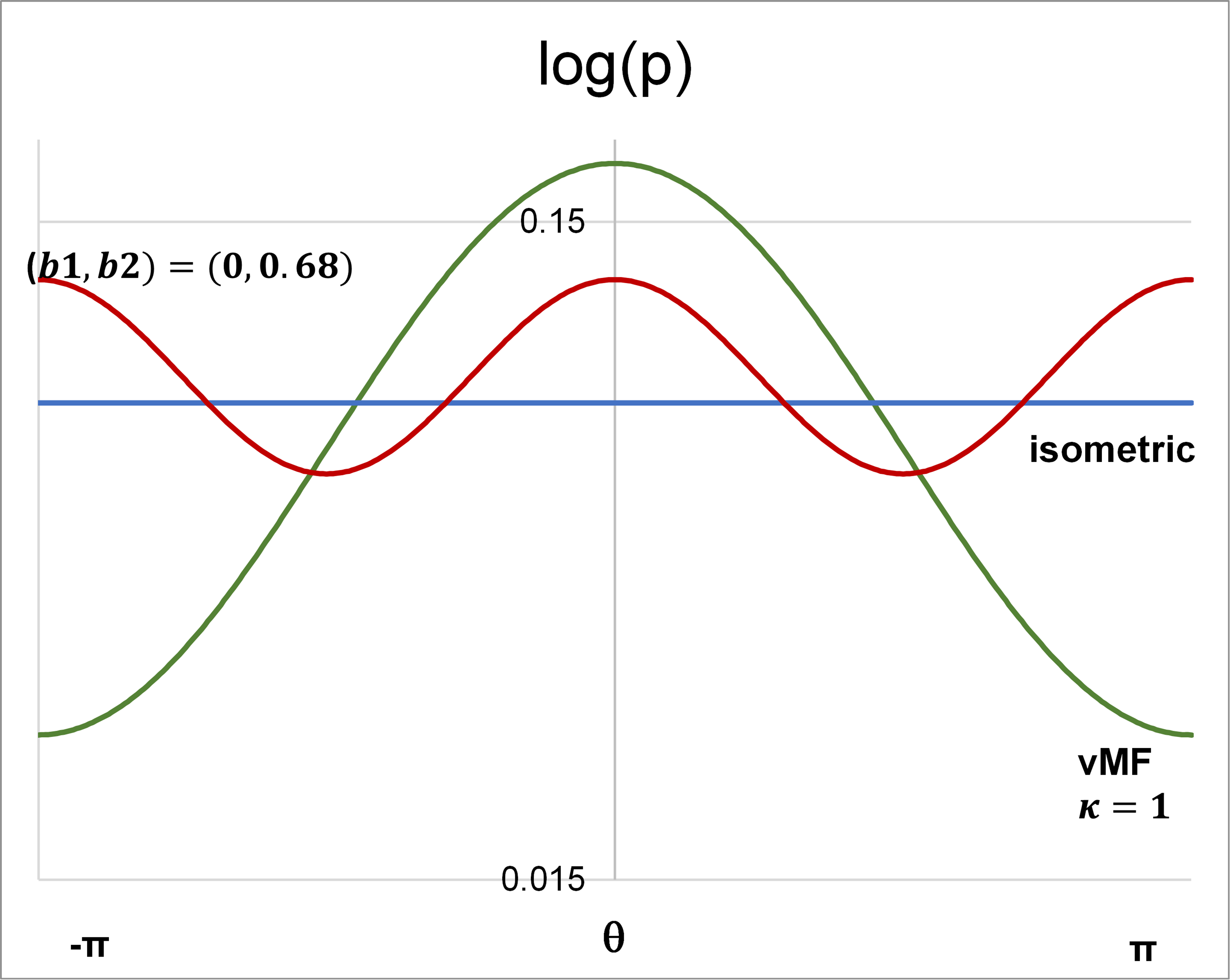}
        \caption{Examples of proposed phase functions of zero-degree, one-degree, and two-degree. }
        \figlabel{pf_examples}
    \end{figure}

    
Although the proposed phase function can theoretically guarantee to fit any positive smooth phase function with a high polynomial degree, the question arises as to how large is sufficient. The question is partially answered in~\cite{Gkioulekas_pf_2013}, where the one-degree exponential phase function may already be useful, and it can be better than HG phase function in a specific range of material. In the next section, we analyze the proposed phase function's representation capability of up to seven degrees and compare it with existing state-of-the-art phase functions.

\subsection{Evaluation with Mie Scattering Theory}
\seclabel{pf_evaluation}
We fit the proposed phase function with those that are predicted by Mie scattering theory~\cite{bohren2008absorption}. There are two prepared datasets of Mie phase functions. The first dataset includes Mie phase functions for media of the identical size particles, which is denoted as \emph{mono-dispersion}. The second dataset includes bulk phase functions for media of particles with a log-normal distribution of size, which is denoted as \emph{poly-dispersion}. We used a publicly-available toolbox~\cite{MieSimulatorGUI} to prepare the two datasets. Light frequency was fixed at 600 nm. In the case of the \emph{poly-dispersion} set, the standard deviation of the particle diameters is 10\% of the mean particle diameter. The details of size settings and asymmetry properties for the simulated media are shown in~\tabref{Mie_exp_size}. We compare the fitting errors between the proposed phase function and several state-of-the-art ones including the polynomial phase functions~\cite{Sharma1998}, described in~\equref{pf_poly} with $P_i=\mu^i $ and $N \in \{7,5,3\} $, HG~\cite{Henyey1941}, and Two-term HG~\cite{KATTAWAR1975} phase functions. The proposed phase functions are with $Q_i=\mu^i $ and $M \in \{7,5,3,1\} $. These phase functions are denoted as \emph{polynomial 7}, \emph{polynomial 5}, \emph{polynomial 3}, \emph{HG}, \emph{Two-term HG}, \emph{exponential 7}, \emph{exponential 5}, \emph{exponential 3}, and \emph{exponential 1}, respectively. It is noted that the \emph{exponential 1} is the same with vMF phase function~\cite{Gkioulekas_pf_2013}. To efficiently handle the extremely high forward scattering, we fit the logarithm of the phase functions and use the sum of absolute difference (SAD) as the fitting error. 

The results for the two datasets are shown in~\figref{fitting_graph}, and some fitting examples are shown in~\figref{fitting_examples}. There are missing results for polynomial phase functions at the left sides of \figref{fitting_graph} a) and b) and in \figref{fitting_examples} b) and c). This is because they exhibited negative values, and the fitting could not be performed successfully, especially for strongly forward scattering cases. Otherwise, the polynomial phase function can fit the Mie data well. For media with small particles, most phase functions can fit the data quite well with small SAD, except that \emph{HG} and \emph{polynomial 1} (or vMF) phase functions do not perform well because they are single-lope phase functions while small particles exhibit both forward and backward scattering. Overall, \emph{HG} or \emph{Two-term HG} does not work well with strong forward scattering media. Meanwhile, the proposed phase functions from three-degree can fit the Mie phase functions of a wide range of particle size better than polynomial, HG, and Two-term HG phase functions do. This demonstrates the advantage of the proposed phase function.
    \begin{table*}[h]
        \centering
        \begin{tabular}{l c c c c c c c c c c c c c }
        \hline
         Diameter (micrometer)	&30	&20	&15	&10	&5	&3	&2	&1	&0.5	&0.3	&0.2	&0.1	&0.01	\\
Mono-dispersion asymmetry	&0.9905	&0.9929	&0.9923	&0.9963	&0.9953	&0.9907	&0.982	&0.95	&0.84	&0.65	&0.33	&0.08	&0.000789	\\
Poly-dispersion asymmetry	&0.9939	&0.9909	&0.9911	&0.9917	&0.9956	&0.9932	&0.985	&0.95	&0.84	&0.65	&0.33	&0.08	&0.000789	\\
        \hline
        \end{tabular}
        \caption{Particle size setting and asymmetry (computed by \equref{g}) of the simulated media.}
        \tablabel{Mie_exp_size}
    \end{table*}

    \begin{figure*}[h]
        \includegraphics[width=\textwidth]{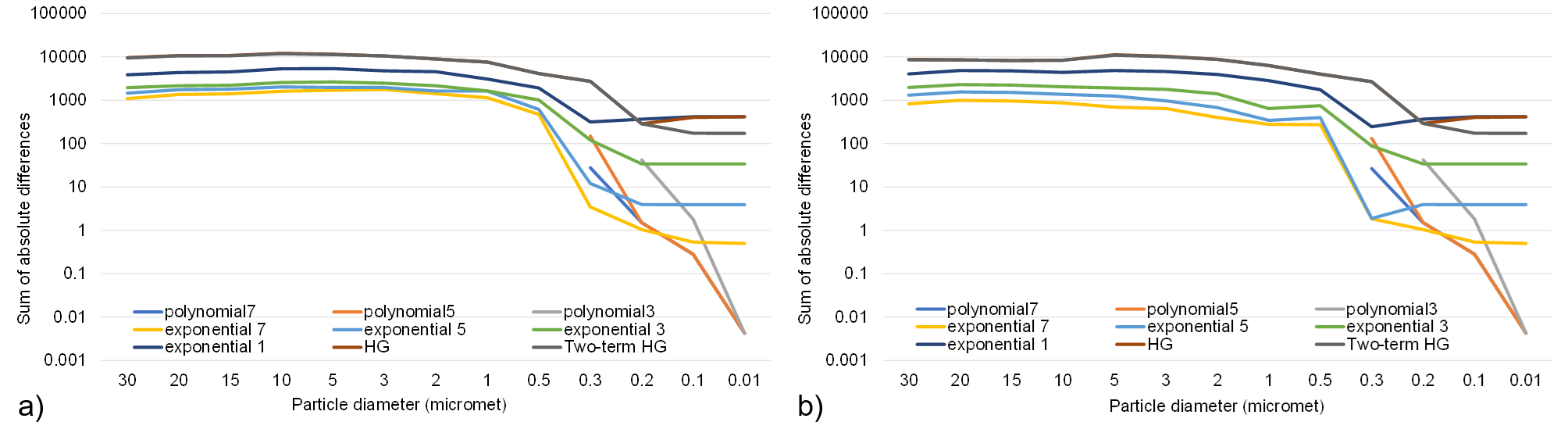}
        \caption{Fitting with Mie scattering phase functions: a) for \emph{mono-dispersion} and b) for \emph{poly-dispersion} datasets.}
        \figlabel{fitting_graph}
    \end{figure*}
    \begin{figure*}
         \includegraphics[width=\textwidth]{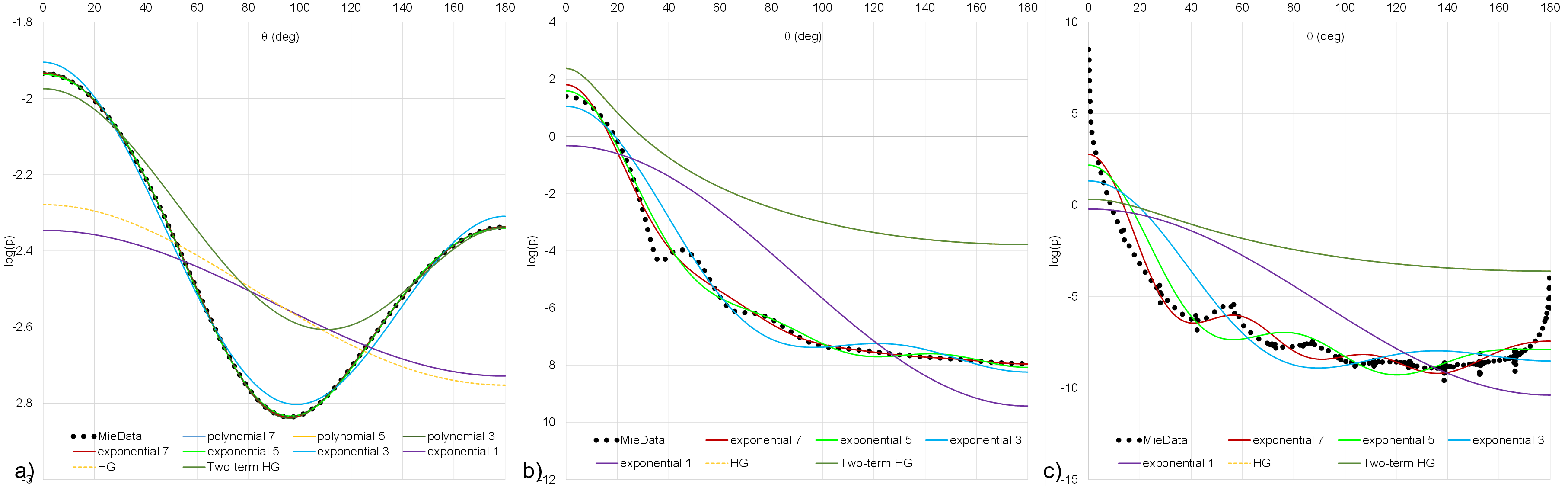}
        \caption{Examples of fitting with Mie scattering phase functions of \emph{poly-dispersion} dataset for mean diameter of a) 0.1 micrometers, b) 1 micrometers, and c) 10 micrometers.}
        \figlabel{fitting_examples}
    \end{figure*}
\section{Inverse Rendering with Exponential Phase Function}
    \begin{figure}
        \centering
        \includegraphics[width=\columnwidth]{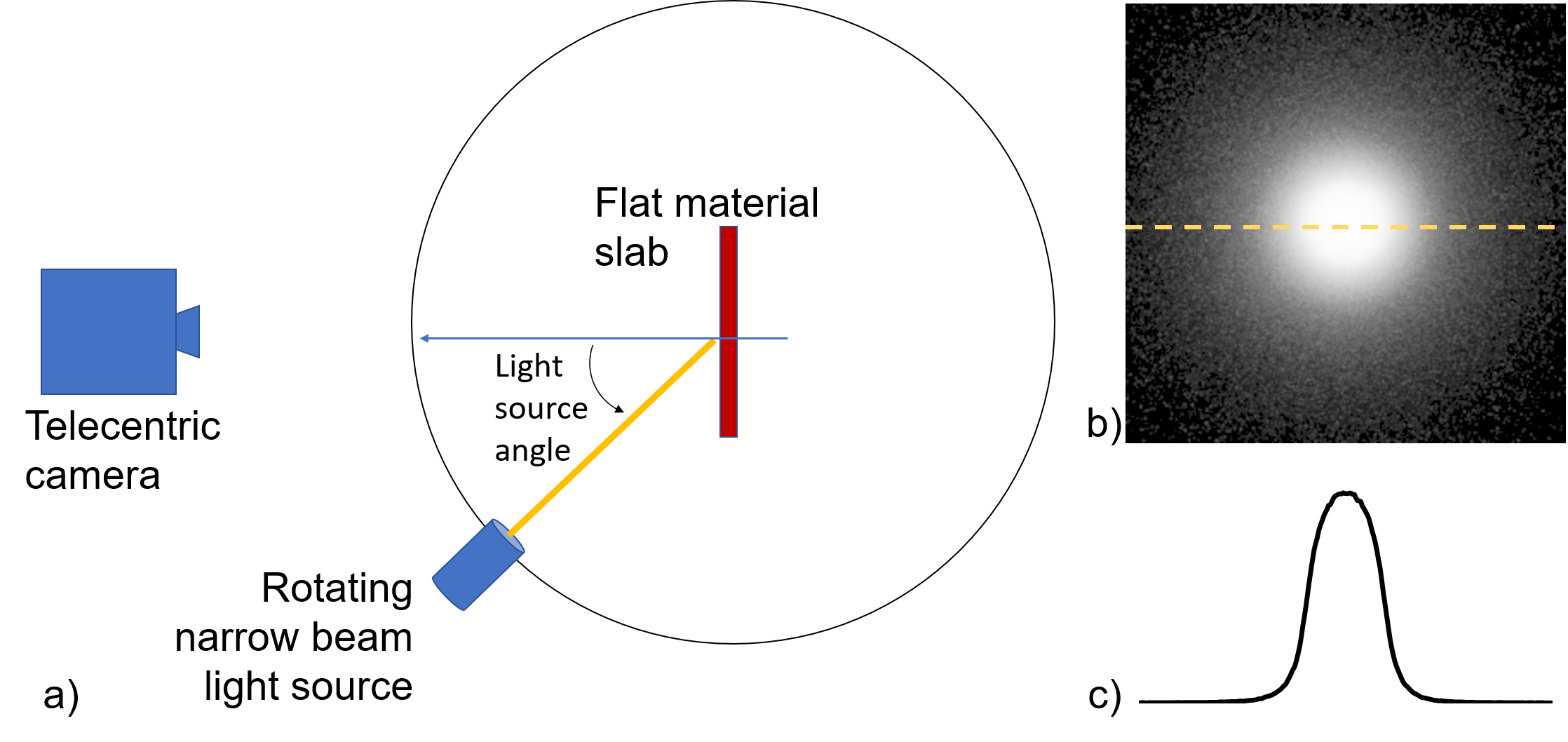}
        \caption{a) Experimental setup includes a telecentric camera, narrow beam light source, and flat thin material slab. The light source can rotate horizontally around the material slab located at the rotation axis. b) An example of rendered image and the horizontal (yellow) line of pixels used for the estimation and c) the intensity profile of the pixel line. }
        \figlabel{scene_setting}
    \end{figure}
We demonstrate an application of the proposed phase function for inverse rendering with a physics-based differential renderer. The problem-setting is described as follows. We estimate scattering parameters (extinction and scattering coefficients and phase function) from a set of captured images of a simple and geometry-known slab of homogeneous translucent material and a known beam light source and its directions. The scene-setting is shown in~\figref{scene_setting}a, which is inspired by~\cite{Gkioulekas_SIGG_2013}. The parameters are solved by the least-squares through an analysis-by-synthesis approach:
    \begin{equation}
        \sigma_t^*, \sigma_s^*, \bm{\phi}_p^* = \underset{\sigma_t,\sigma_s,\bm{\phi}_p}{\arg\min} \sum_{i=1}^{K}\big{(}I_{\bm{l}_i} - I(\sigma_t,\sigma_s,\bm{\phi}_p, \bm{l}_i) \big{)}^2,
        \equlabel{L2_opt}
    \end{equation}
    where $\sigma_t,\sigma_s,\bm{\phi}_p$ are extinction, scattering coefficients, and phase function parameters, respectively; $\bm{l}_i$ is a beam light source direction; $I_{\bm{l}_i}$ is captured image when light source is at $\bm{l}_i$; and $I(\sigma_t,\sigma_s,\bm{\phi}_p, \bm{l}_i)$ is synthesized image rendered by a physics-based renderer. To ignore the light source intensity difference between real and synthesized scenes, images are normalized so that the average of image intensity of the whole image set is 1. 
    
The experiments in~\secref{pf_evaluation} encourage us to use the three-degree proposed phase function with the base polynomial $Q_i=\mu^i$ for this application and thus $\bm{\phi}_p=(b_1,b_2,b_3)$. We take the Two-term HG phase function for the reference since it also has three parameters. We implemented the proposed phase function and Two-term HG phase functions on top of a physics-based differentiable renderer provided by Che et al.~\cite{Che_ICCP2020}, which is based on Mitsuba 0.5~\cite{Mitsuba05}. An optimization routine was implemented that handled both phase functions similarly.
To reduce the computational cost for rendering, we use only a horizontal line of pixels on the symmetric line of scattering spot from each captured image and also synthesize just a single line, as illustrated in~\figref{scene_setting}b and \figref{scene_setting}c. To further reduce the computational cost, we start rendering $I(\sigma_t,\sigma_s,\bm{\phi}_p, \bm{l}_i)$ with a low sample-per-pixel (i.e., 128) and gradually increase this number whenever the loss get smaller and stable. 
    
However, the least square optimization~\equref{L2_opt} faces a problem of very different dynamic ranges of captured images. It tends to overfit the extremely high intensity even though both the peak and tail of a scattering spot are equally important to our solution. In such a situation, log-intensity technique~\cite{Oxholm_PAMI15} is usually employed. However, the log-intensity technique amplifies the noise of the dark pixels that are associated with a very low signal-to-noise rate. We eventually use a log-intensity of the image after adding some positive bias $\delta$ and the optimization~\equref{L2_opt} becomes:
    \begin{equation}
        \sigma_t^*, \sigma_s^*, \bm{\phi}_p^* = \underset{\sigma_t,\sigma_s,\bm{\phi}_p}{\arg\min} \sum_{i=1}^{K}\big{(}   \log(I_{\bm{l}_i}+\delta) - \log(I(\sigma_t,\sigma_s,\bm{\phi}_p, \bm{l}_i)+\delta)           \big{)}^2.
        \equlabel{log_opt}
    \end{equation}
Since logarithm function is monotonic, the global solutions, if exists, of \equref{L2_opt} and \equref{log_opt} are the same for any $\delta$. There is an additional benefit from the reformulated optimization, described as follows. Now, the optimization can be more robust because the logarithm function can modify the dynamic range of the original intensity by splitting it into two sub-ranges. The lower sub-range is amplified while the upper sub-range is attenuated, illustrated in~\figref{log_technique}a, and hence the optimization focuses more on the lower sub-range. By randomly alternating $\delta$, the optimization can change its focus on the different sub-range of the original intensity and secure a global solution that survives through all these $\delta$s. The transformation of the original intensity with different $\delta$ is illustrated in~\figref{log_technique}.

    \begin{figure}
        \centering
        \includegraphics[width=\columnwidth]{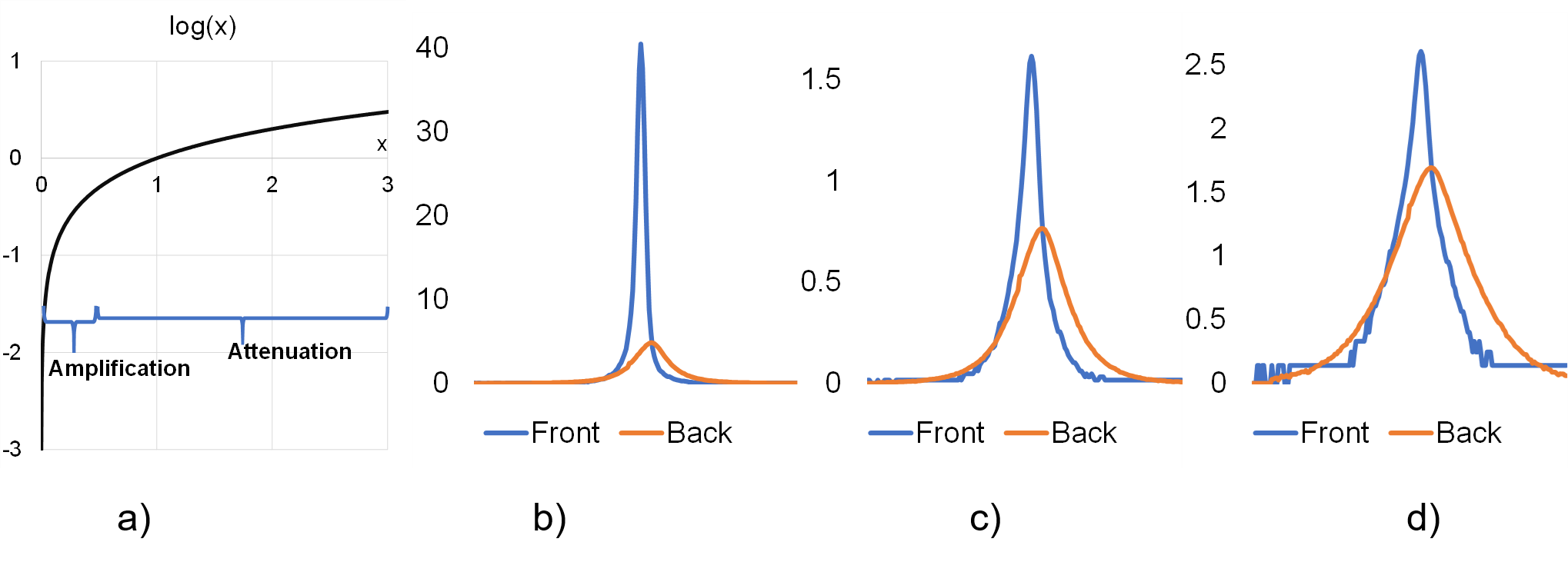}
        \caption{a) Logarithm function splits the original range into 2 sub-ranges, the lower sub-range is amplified and the upper sub-range is attenuated. b) Examples of original real captured intensity profiles for the light source at front and back of the slab. The transformed intensity profiles are shown in c) and d) for $\delta$ = 1 and 0.1, respectively. It is noted that optimization with the original intensity profiles of b) is really difficult because the results would overfit the extremely high intensities of the front profile. While optimization with transformed profiles of d) would overfit the noisy dark intensities.}
        \figlabel{log_technique}
    \end{figure}
    

\section{Experiments}
\subsection{Simulation Experiment}
We selected Mie phase functions from the \emph{poly-dispersion} dataset in \secref{pf_evaluation} with the particle diameters of 0.5, 0.3, 0.2, and 0.01 micrometers, $\sigma_t$ = 2,6, and fixed the volume albedo at 0.9 to render a dataset. The simulated slab has a thickness of 1 mm. For the stronger forward scattering phase function with larger particle size or optically thinner media with a lower $\sigma_t$, the dynamic ranges of images between the front and back-lighting are too different, and the optimization does not work well. The optimization also did not work well for optically thicker media with stronger $\sigma_t$. For each set of $\sigma_t$, albedo, and phase function, we rendered images in ten light source directions (i.e., five images for front-lighting and five images for back-lighting). 

    \begin{figure}
        \centering
        \includegraphics[width=\columnwidth]{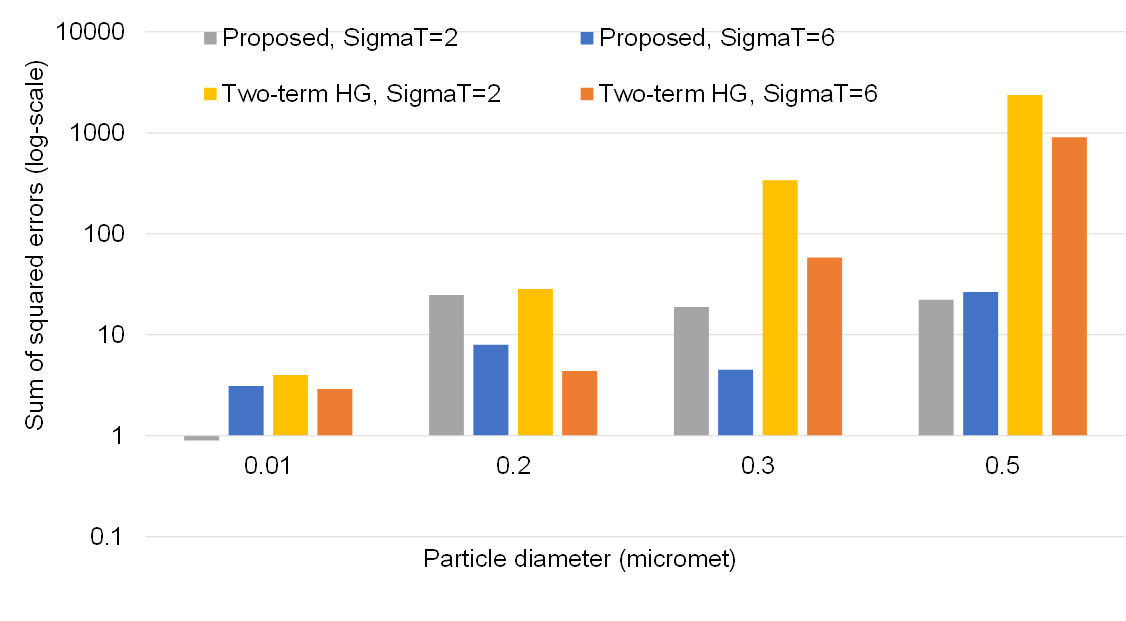}
        \caption{The fitting error for scattering parameter estimation with the proposed and the Two-term HG phase functions. The fitting error is computed by the sum of squared errors between the target and the rendered intensity profiles.}
        \figlabel{simu_fitting_error}
    \end{figure}
    
    \begin{figure}
        \centering
        \includegraphics[width=\columnwidth]{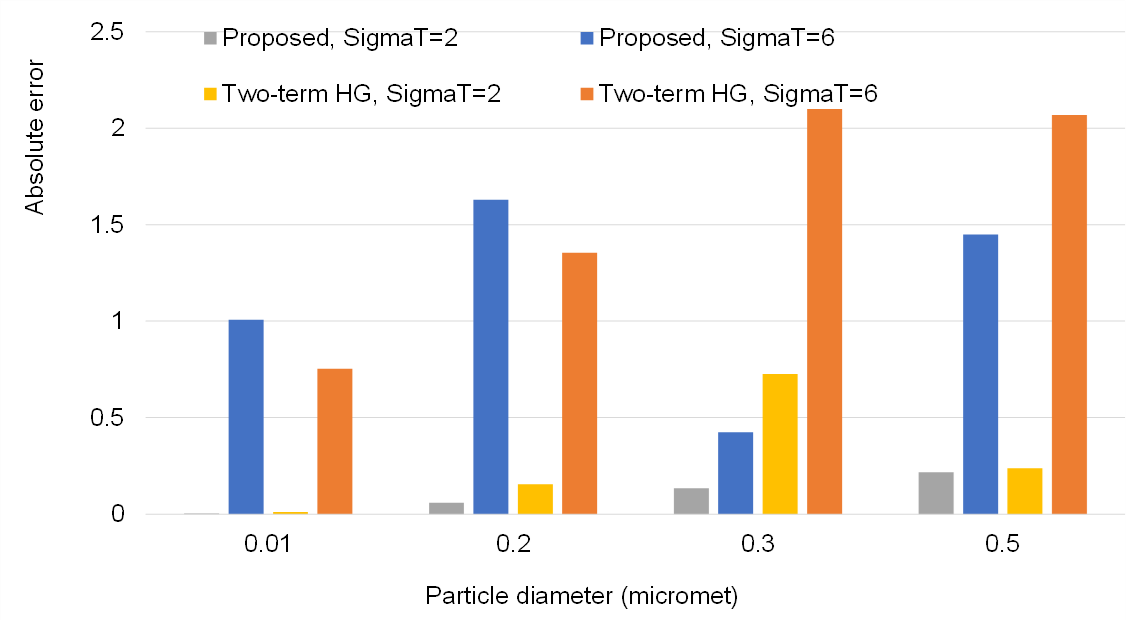}
        \caption{The absolute error between the estimated $\sigma_t$ and the ground-truth.}
        \figlabel{simu_sigmaT_error}
    \end{figure}
    
    \begin{figure}
        \centering
        \includegraphics[width=\columnwidth]{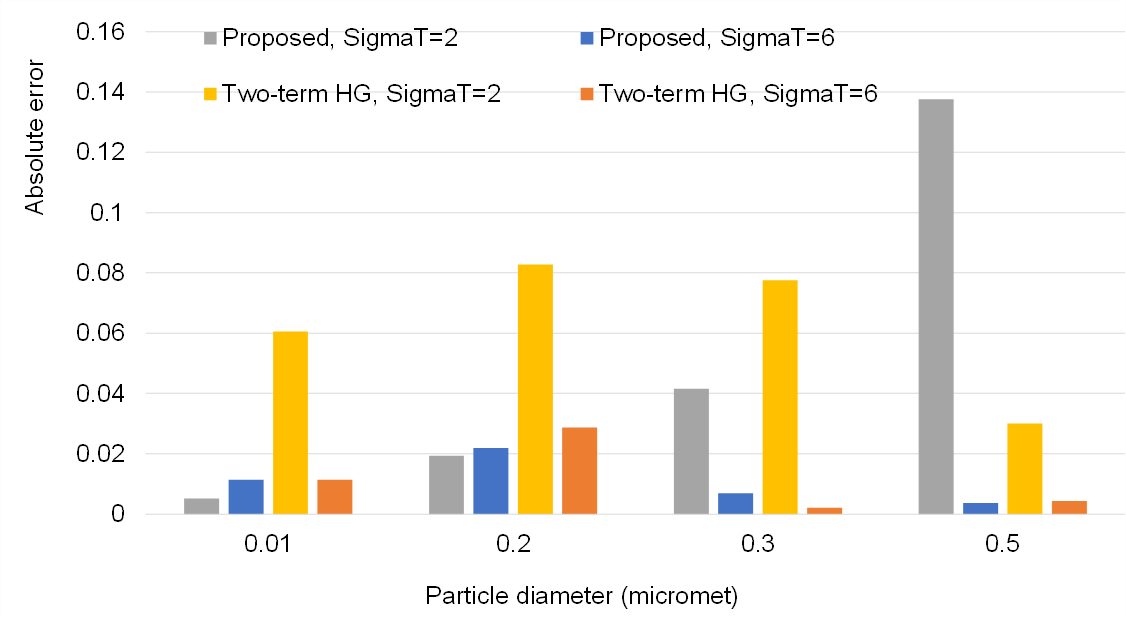}
        \caption{The absolute error between the estimated albedo and the ground-truth.}
        \figlabel{simu_albedo_error}
    \end{figure}
    
    The estimation performances are described in~\figref{simu_fitting_error} for the fitting error, \figref{simu_sigmaT_error} for $\sigma_t$ error, and \figref{simu_albedo_error} for albedo error. The optimization method works well with both phase functions for optically thinner and smaller particle media. The optimization performance is worst if the media is optically thin and contains large particles (with strong forward scattering). The overall performance with the proposed phase function is better than that with the Two-term HG phase function with lower fitting errors, higher $sigma_t$ and albedo accuracy. However, there are cases that the optimization with the proposed function does not work well such as for optically thin and strongly-forward scattering media, as shown on the right side of~\figref{simu_albedo_error}.
    
\subsection{Real-world Experiment}
    \begin{figure}
        \centering
        \includegraphics[width=\columnwidth]{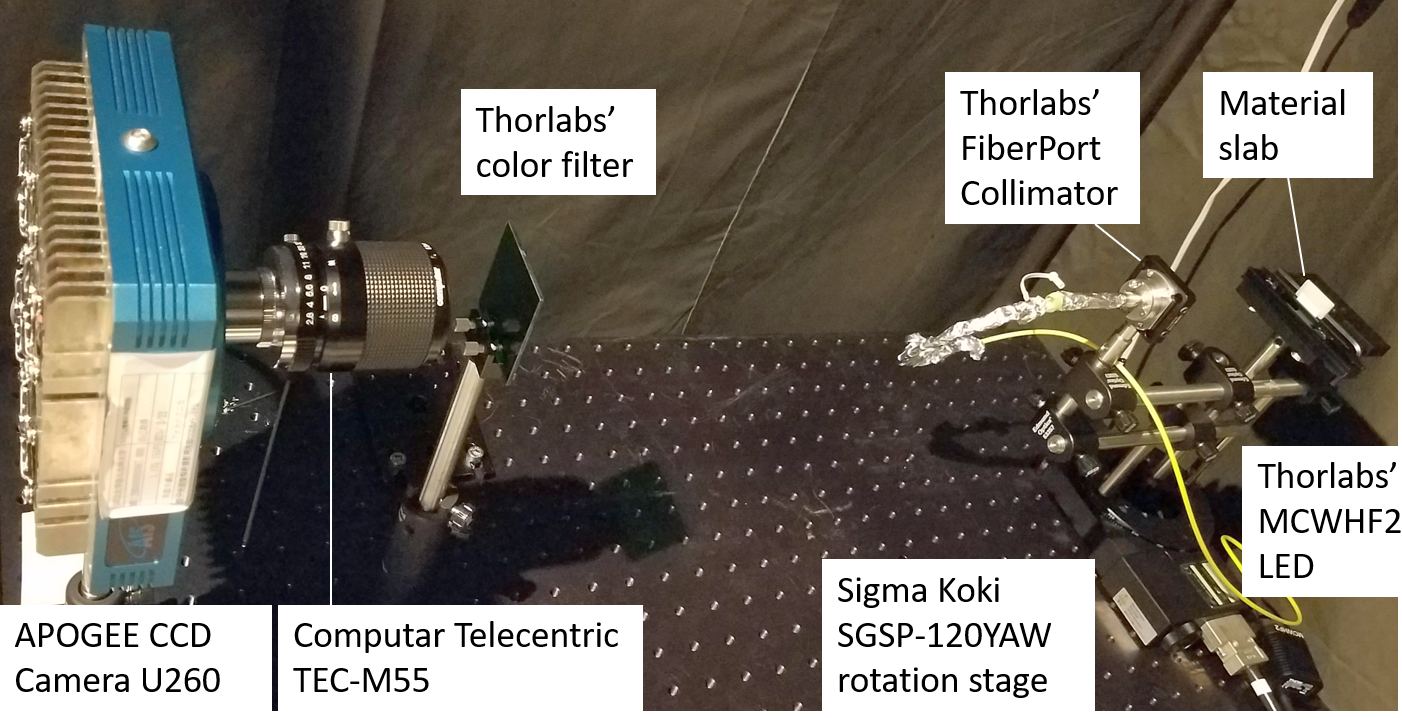}
        \caption{Real-world experiment setup includes an Apogee ALTA U260 CCD camera with a resolution of 512x512, a Computar TEC-M55 telecentric lens, a Thorlabs' MCWHF2 white LED light source, a Thorlabs' fiberport collimator which was mounted on top of a Sigma Koki SGSP-120YAW rotation stage. We used three color filters (red, green, blue) for preparing the real-world experimental dataset.}
        \figlabel{real_exp_setup}
    \end{figure}
    \begin{figure}
        \centering
        \includegraphics[width=\columnwidth]{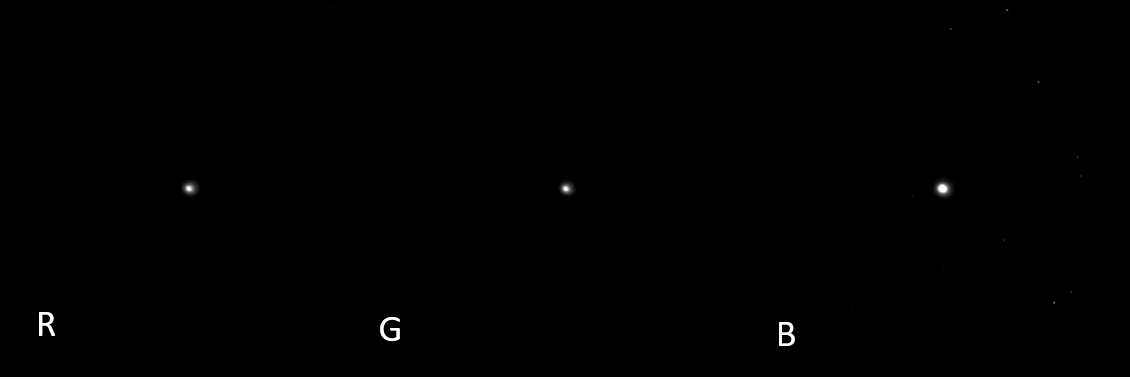}
        \caption{The light source patterns are captured separately for three color filters.}
        \figlabel{real_exp_LS_pattern}
    \end{figure}
    
The real-world experiment was conducted with an imaging system shown in~\figref{real_exp_setup}. The light source pattern with each color filter was captured, shown in~\figref{real_exp_LS_pattern}, and they were used for the differential renderer. The target material objects include a 1.2 mm yellow wax, a 1.2 mm white rubber eraser, a 1.2 mm dry yellow clay, and a 2.5 mm cheese slab. We captured and constructed ten high-dynamic-range images for red, green, and blue color filters in ten light directions. Therefore, we have twelve subsets of images. We conducted the experiments for these subsets independently. Because there is no ground-truth phase function for this dataset, we evaluate the performance of scattering parameters by the fitting errors between the captured and rendered images. The results are shown in~\figref{Real_fitting_errors}, and an example of fitting for the blue channel of the yellow wax is shown in~\figref{Real_fitting_wax}. The rendered images of the materials and the actual material slabs are shown in~\figref{teaser}. It is clear that with the proposed phase function, the scattering parameters can produce much better fitting performances than those with the Two-term HG phase function. For the Two-term HG, most of the errors come from the fitting with the peak intensities of the data profiles as illustrated in~\figref{Real_fitting_wax}. The real-world experiment proves that the proposed phase function can well describe the real-world scattering phase function.
    \begin{figure*}[ht]
        \centering
        \includegraphics[width=2\columnwidth]{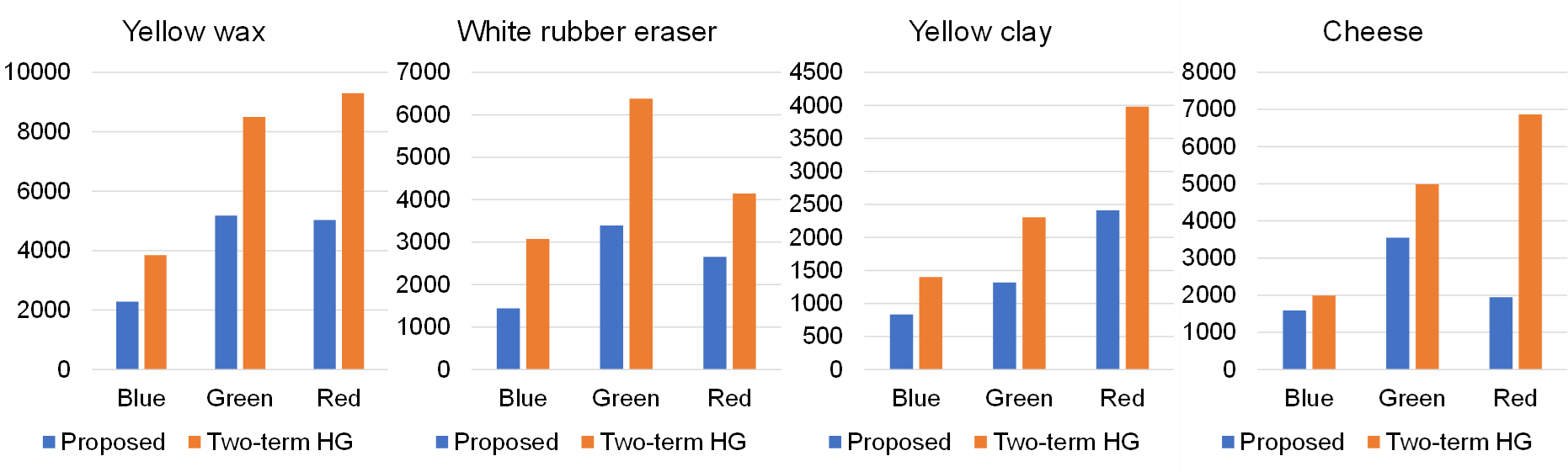}
        \caption{The fitting errors of the estimation methods using the proposed and Two-term HG phase functions for the four material slabs and three color channels. The fitting error is computed by the sum of squared errors between the captured and rendered intensity profiles.}
        \figlabel{Real_fitting_errors}
    \end{figure*}
    
    \begin{figure*}
        \centering
        \includegraphics[width=\textwidth]{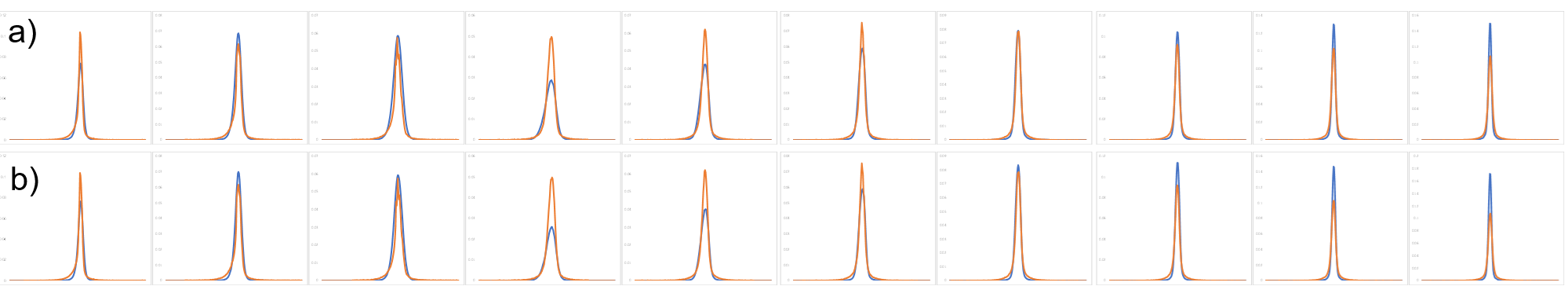}
        \caption{An example of fitting the yellow wax for the blue color channel for a) proposed phase function and b) Two-term HG phase function. The light source moved from the front to back for the images from left to right. Captured profiles are shown in red while the rendered profiles are shown in blue.}
        \figlabel{Real_fitting_wax}
    \end{figure*}
    
\section{Conclusions and Future Works}
We present the novel exponential phase function, which has the ability to theoretically represent most existing empirical phase functions and an application for studying the translucent material. The fitting capability of the proposed phase function enables it to tightly fit the phase functions predicted by Mie scattering theory and the real-world data. The initial study has positively confirmed the capability of the proposed phase function.

In the current study, we just employ a few parameters. However, we believe that the capability of the proposed function can be better with more parameters being used. We plan to study more real-world materials, including both solid and liquid materials, to further confirm its advantages.

\section*{Acknowledgement}
We acknowledge open-source software resources offered by the Virtual Photonics Technology Initiative, at the Beckman Laser Institute, University of California, Irvine.

\bibliographystyle{ACM-Reference-Format}
\bibliography{myref}
\end{document}